\documentclass[letterpaper, 10pt, conference]{ieeeconf}      

\IEEEoverridecommandlockouts                              

\usepackage{graphics} 
\usepackage{epsfig} 
\usepackage{times} 
\usepackage{amsmath} 
\usepackage{amssymb}  
\usepackage{hyperref}
\usepackage{float}
\usepackage[table]{xcolor}
\usepackage{multirow}
\usepackage{makecell}
\usepackage{diagbox}

\newcommand\mdoubleplus{\mathbin{+\mkern-10mu+}}

\def\BibTeX{{\rm B\kern-.05em{\sc i\kern-.025em b}\kern-.08em
    T\kern-.1667em\lower.7ex\hbox{E}\kern-.125emX}}
\makeatletter
\DeclareRobustCommand\onedot{\futurelet\@let@token\@onedot}
\def\@onedot{\ifx\@let@token.\else.\null\fi\xspace}

\makeatother

\def\re{\mathbb{R}}

\title{\LARGE \bf Strength in Diversity: Multi-Branch Representation Learning for Vehicle Re-Identification*}

\author{Eurico Almeida$^{1}$, Bruno Silva$^{1}$ and Jorge Batista$^{1,2}$
\thanks{*This work was supported by A-to-Be - Mobility Technology, S.A and Fundação para a Ciência e a Tecnologia (FCT) under the project UIDP/00048/2020.}
\thanks{$^{1}$Eurico Almeida, Bruno Silva and Jorge Batista are with the Institute of Systems and Robotics,
        University of Coimbra, 3030-219 Coimbra, Portugal
        ({\tt\small eurico.almeida, bsilva, batista@isr.uc.pt})} %
\thanks{$^{2}$Jorge Batista is with the Department of Electrical and Computers Engineering, Faculty of Science and Technology, University of Coimbra, 3030-290 Coimbra, Portugal
        ({\tt\small batista@deec.uc.pt})}%
}

\begin{document}

\maketitle
\thispagestyle{empty}
\pagestyle{empty}

\begin{abstract}

This paper presents an efficient and lightweight multi-branch deep architecture to improve vehicle re-identification (V-ReID). While most V-ReID work uses a combination of complex multi-branch architectures to extract robust and diversified embeddings towards re-identification, we advocate that simple and lightweight architectures can be designed to fulfill the Re-ID task without compromising performance.

We propose a combination of Grouped-convolution and Loss-Branch-Split strategies to design a multi-branch architecture that improve feature diversity and feature discriminability. 
We combine a ResNet50 global branch architecture with a BotNet self-attention branch architecture, both designed within a Loss-Branch-Split (LBS) strategy. We argue that specialized loss-branch-splitting helps to improve re-identification tasks by generating specialized re-identification features. A lightweight solution using grouped convolution is also proposed to mimic the learning of loss-splitting into multiple embeddings while significantly reducing the model size.
In addition, we designed an improved solution to leverage additional metadata, such as camera ID and pose information, that uses 97\% less parameters, further improving re-identification performance.

In comparison to state-of-the-art (SoTA) methods, our approach outperforms competing solutions in Veri-776 by achieving 85.6\% mAP and 97.7\% CMC1 and obtains competitive results in Veri-Wild with 88.1\% mAP and 96.3\% CMC1. Overall, our work provides important insights into improving vehicle re-identification and presents a strong basis for other retrieval tasks. Our code is available at the \href{https://github.com/videturfortuna/vehicle_reid_itsc2023}{link}.
\end{abstract}

\section{Introduction}

\begin{figure}[t]
\centerline{{\includegraphics[width=\linewidth]{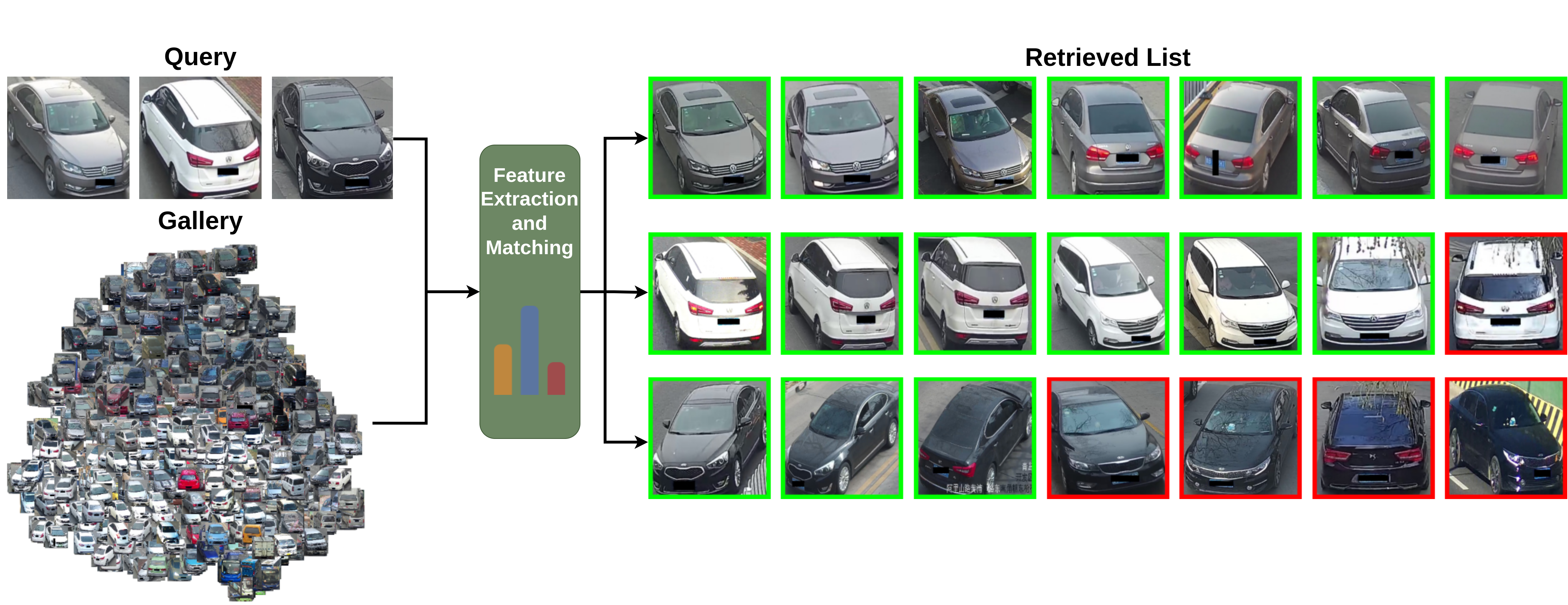}}}
\caption{Three examples of vehicle re-identification, along with their top seven retrieved matches.}
\label{fig:example_reid}
\end{figure}

The task of vehicle re-identification (V-ReID) involves matching images of the same vehicle from a large gallery set when presented with a query image, as depicted in Fig.~\ref{fig:example_reid}. V-ReID has significant practical applications in fields such as self-driving vehicles, smart cities, and traffic monitoring, and can provide a reliable alternative to other methods such as sensors and license plate recognition. 

Deep Neural Networks (DNNs), particularly Convolutional Neural Networks (CNNs), have revolutionised computer vision tasks by significantly outperforming traditional methods that rely on hand-crafted features.
However, V-ReID with CNNs still poses significant challenges due to inter-class similarities and intra-class discrepancies, such as different vehicles of the same model and color or the same vehicle having an entirely different appearance when observed from a different angle.

To address these challenges, we propose an approach that generates more robust image features by using multiple representations learned differently. Our model concatenates diverse global representations learned by multiple branches with different architectures, losses  or applied in different channel groups to achieve a superior outcome. 
We advocate that using multiple branch learned with dedicated losses, different architectures or acting on different channel groups per branch offer several advantages for V-ReID tasks:
\begin{itemize}
\item \textbf{Improved feature diversity}: Each branch of the backbone network learns different features, and using multiple branches can increase the diversity of features that are captured. This can lead to better discrimination between different vehicles and improve the accuracy of the re-identification system.
\item \textbf{Reduced overfitting}: When using a single branch for feature extraction, there is a risk of overfitting to the training data. However, by using multiple branches, the model can learn multiple representations of the data, which can help reduce overfitting and improve generalisation to new data.
\item \textbf{Increased robustness}: Multiple branches can help the model become more robust to changes in the input data. The community has proven that, by using multiple branches architectures designed distinctively or learned with different data and losses, lead to improved performance and better feature representation, apart from more efficient training and improved scalability, making it a powerful approach for object re-identification tasks.

\end{itemize}
Instead of relying on part-based or attribute-based techniques which also require multiple branches or additional models, our method employs an ensemble of richer and more expressive global embeddings that increase flexibility, improve transferability, and enhance robustness, ultimately leading to better performance in V-ReID tasks. 
Our research consist in four contributions:
\begin{itemize}
    \item Use of a Loss-Branch-Split (LBS) architecture that aggregates different architectures and losses per branch to generate diverse global embeddings. 
    \item Make use of a self-attention branch to capture local dependencies between vehicle's discriminant parts.
    \item Use grouped convolutions to obtain a lightweight model of the LBS architecture, overcoming the increased complexity of the LBS architecture.

    \item Efficient leveraging of additional metadata information with CNN, namely camera ID and vehicle's pose.
\end{itemize}

\begin{figure*}[hbt!]
\centerline{{\includegraphics[width=0.8\linewidth]{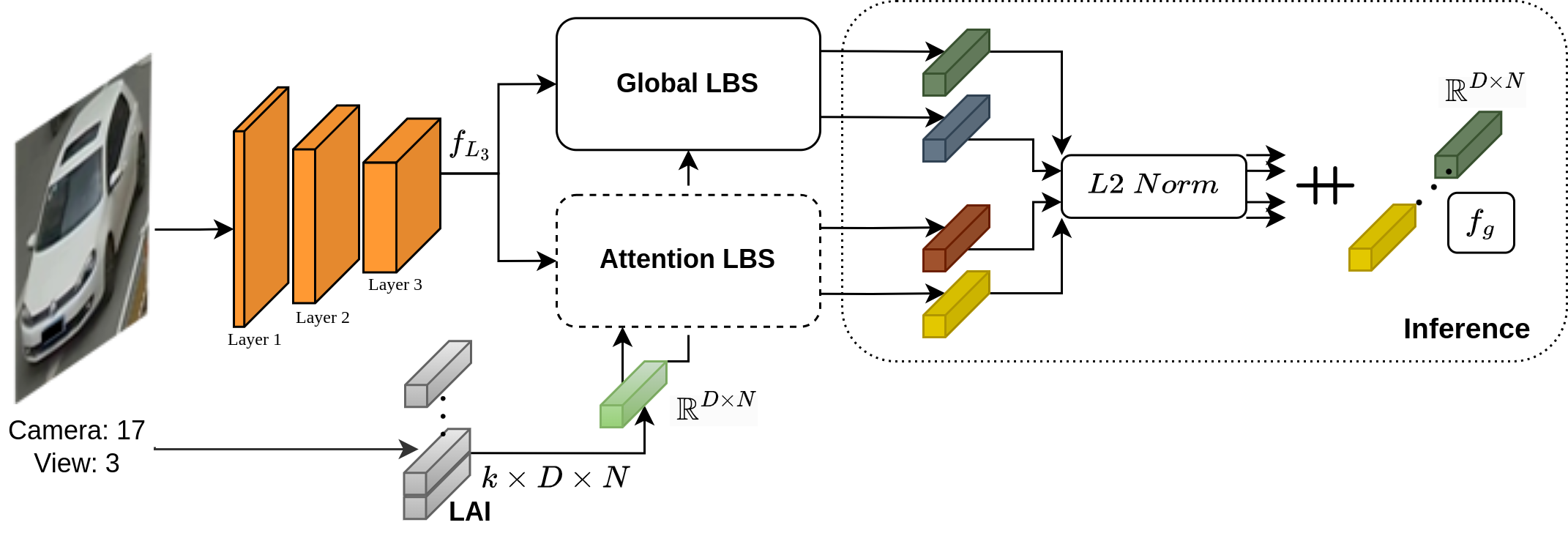}}}
\caption{Diagram of the proposed MBR architecture. In this figure $\mdoubleplus$ denotes concatenation.} 
\label{fig:diagrama}
\end{figure*}

In summary, our proposed improvements address the challenges of V-ReID with CNNs by incorporating a number of novel and efficient strategies. Resourcing to architectural modifications and using different losses in each branch to extract diversified vehicle representations, and by efficiently leveraging additional metadata, our solution achieved improved accuracy and performance in re-identification tasks.
\section{Related Work}
The vehicle re-identification problem has been tackled in various ways, including part-based \cite{meng2020parsing, he2019part, lee2022multiple, suprem2020looking}, attribute-based \cite{quispe2021attributenet, sun2020CFVMNet, article}, local-based \cite{wang2018learning}, self-supervised \cite{khorramshahi2020devil, khorramshahi2022scalable, li2021self}, and global methods \cite{he2021transreid, zhao2021heterogeneous, he2020fastreid}. As ours, many of these approaches use multiple branches on the last stage or models to extract additional information to the global representation and generate more robust features. However, the ratio of accuracy gain over the computational cost is sometimes not good enough, limiting the use of these approaches for real-time applications.

Part-based approaches involve capturing specific parts of the vehicle to aid in vehicle's re-identification. For example, PEVEN \cite{meng2020parsing} uses a segmentation model to identify visible parts of the vehicle and generate embeddings specific to those parts, such as the front, rear, side, and top of the vehicle. He \textit{et al.} \cite{he2019part} uses detection models to identify finer-detail parts like the windshield, brand logo, and lights to generate part-based embeddings. Lee \textit{et al.} \cite{lee2022multiple} also uses multiple representation by aggregating a global embedding to n-1 attention based part features generated by spatial and channel attention. 
Other part-based strategies, as Wang \textit{et al.} \cite{wang2018learning} utilise multi-branches to extract different granularities of local features by splitting the final features horizontally $n$ times at each branch. These models often become computational expensive by requiring additional models or branches to detect the specific parts.  

Attribute-based methods \cite{quispe2021attributenet, sun2020CFVMNet, article} also use multiple branches in the later stages of the backbone to extract embeddings beside the global representation related to each attribute or use attention to extract attribute embeddings from the global features. 

Self-supervised methods attempt to improve representations using input data to generate their own supervision. For instance, Khorramshahi \textit{et al.} \cite{khorramshahi2020devil} uses a variational autoencoder to obtain attention in uncommon parts where it fails to reconstruct, while in \cite{khorramshahi2022scalable} a teacher-student framework is used where the teacher is a momentum encoder of the student and each are fed with different crops of the image like DINO \cite{caron2021emerging}. Li \textit{et al.} \cite{li2021self} uses an extra ResNet-18 model to obtain "landmarks" through different rotations capturing an attention map for learning the global feature.

Other methods aim to improve global embeddings, such as through different backbones \cite{he2021transreid}, multi-branches \cite{sun2020CFVMNet, meng2020parsing, quispe2021attributenet, article, lee2022multiple}, graph-based \cite{zhao2021heterogeneous}, and post-processing techniques \cite{zhong2017re}. 
Zhai \textit{et al.} \cite{Zhai_2019_CVPR_Workshops} proposes a multi-branch approach in the same line as the one proposed. Contrary to ours, they adopt a re-id framework featured by channel grouping and multi-branch strategy, dividing global feature vector into multiple channel groups and learning the discriminative channel group features by multi-branch classification layers that are uniquely driven by a cross-entropy classification loss.
He \textit{et al.} \cite{he2021transreid} uses additional information as input to improve results. In our work, we design an identical strategy adapted for CNN's with less than 3\% of their computational cost.

Our work also uses a multi-branch architecture, but instead of extracting parts, attributes, or local features, we extract globally diversified features using different architectures and applying distinctive loss learning strategies at each branch.
We advocate that the classification and metric losses commonly used jointly in ReID tasks can provide more discriminative embeddings when considered independently in specialized branches. Further, with grouped convolutions, we allow LBS and different architectures to operate each branch in different channel groups while diminishing the size of the model.
We demonstrate significant performance and accuracy improvements with our simple yet powerful architectural modifications compared to competing similarly sized models.

\section{Methodology} 
The methodology section describes the approach taken to address the vehicle re-identification task. The method combines multiple global descriptors that are learned using a multi-branch architecture. The architecture is split into $N$ branches, each generating a distinctive global embedding. Each branch has a specialized task and gets a particular type of feature extraction leading to an overall more robust and informative representation.

To accomplish such representation we adapted a ResNet50 deep architecture to accommodate different losses per branch and incorporate a transformer branch \cite{srinivas2021bottleneck} to retrieve diversified features using self-attention. 
We further leverage additional metadata similar to \cite{he2021transreid} but adapted to CNN's. A general representation of our proposed architecture is shown in Fig. \ref{fig:diagrama}.

\subsection{Multi-branch Global Embeddings}


The proposed architecture is specifically designed to extract multiple global embeddings by exploring a multi-branch architecture that leverages branch-specific architecture models and branch-loss specialized training (MBR).

As illustrated in Fig.~\ref{fig:diagrama}, our network structure is composed of two main modules, a global module and an attention module, that can be combined distinctively in a branched architecture in order to extract diversified embeddings. Each branch are lately combined with metadata information in an end-to-end CNN framework, further improving re-identification ability.

The backbone consists of a ResNet50-IBN (Instance Batch Normalization) \cite{pan2018two}.
The global module uses layer 4 of ResNet50 to extract global embeddings, while the attention module uses a BoTNet transformer structure to extract global attention embeddings.

\begin{figure}[tb]
\centerline{{\includegraphics[width=0.8\linewidth]{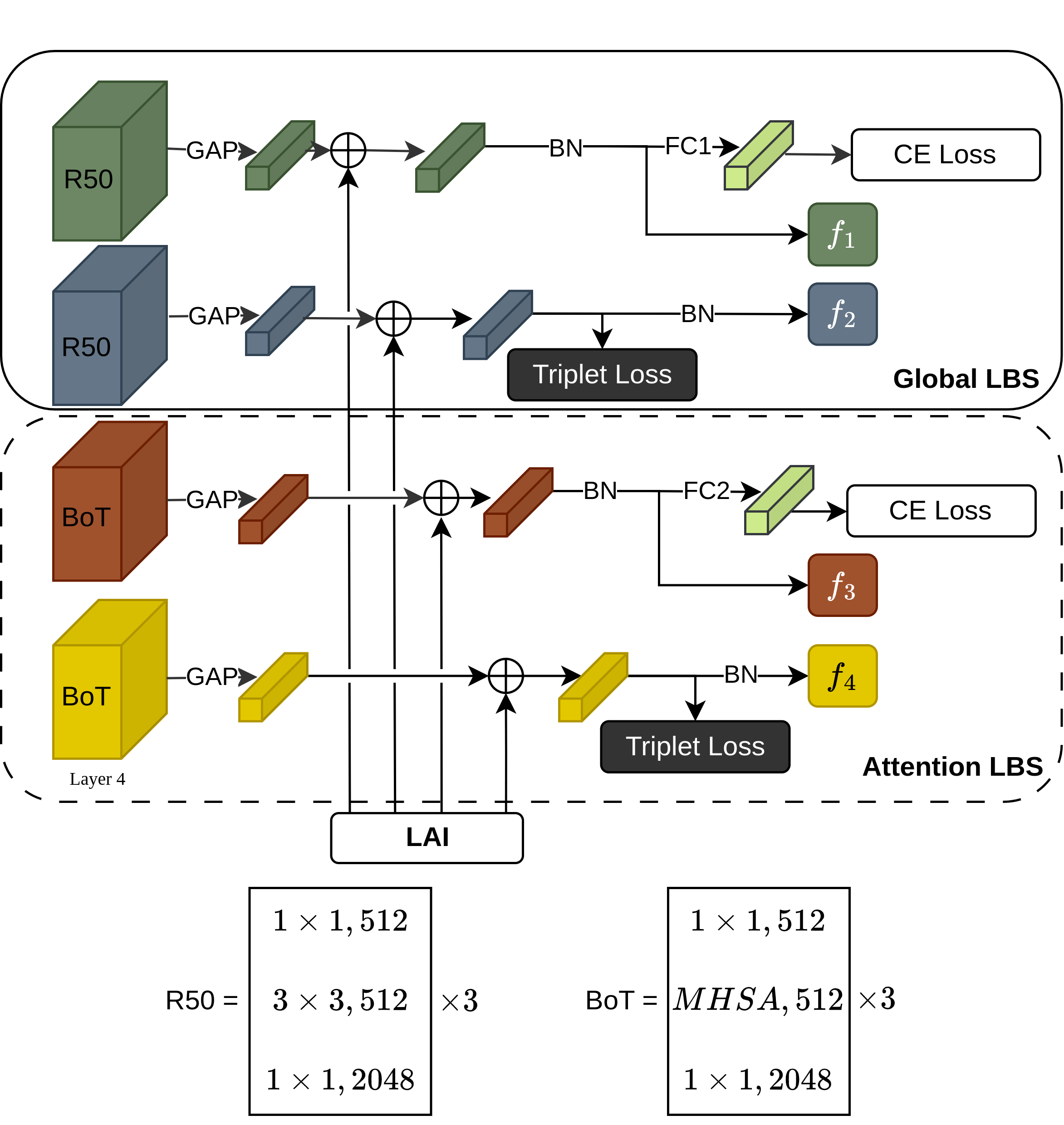}}}
\caption{Global LBS and Attention LBS blocks plus LAI architecture of MBR-4B-LAI. $\oplus$ denotes element wise sum.} 
\label{fig:diag_b_branches}
\end{figure}

We consider distinct branches per loss, one for classification alone and the other only relying on the metric loss. 
By splitting the loss functions, we allow the system to explicitly differentiate between the semantic understanding of identity (classification loss) and the discriminative power of feature embeddings (metric loss).
Fig.~\ref{fig:diag_b_branches} presents the MBR-4B architecture that leverages loss-specific branch training and incorporates two different branches at each architecture module.

The $N$ branches of the architecture share weights until the penultimate layer of the backbone $F_{s_{123}}(x; \theta_{s})$ outputting a feature map $f_{L_3} \in \mathbb{R}^{16 \times 16 \times 1024}$.  Each branch possesses its own stage 4 $F_{N_4}(f_{L_3};\theta_N)$ as represented in the diagram at Fig. \ref{fig:diag_b_branches}. Each branch is generated by,
\begin{equation}
\label{eq:featurebranch}
 f_N(x) = GAP[F_{N_4}(F_{s_{123}}(x; \theta_{s}); \theta_N)],
\end{equation}
where $f_{N}$ is the $N^{th}$ resulting embedding, $F_{N_4}$ is the $N^{th}$ branch layer, $F_{s_{123}}$ is the backbone layers until the last layer, GAP stands for global average pooling, $\theta_s$ are the shared weights, $x$ is a input image, and $\theta_N$ the $N^{th}$ branch weights. 
This architecture outputs an embedding $f_N\in\mathbb{R}^{D}$, where $D=2048$ for each of the $N$ branches used, which are L2 normalized to contribute equally and then concatenated to obtain the final representation $f_g\in \mathbb{R}^{DN}$.

\subsubsection{ResNet50 Global Branch (R50)}
The global branch is designed to capture global representations and simply adapts the ultimate (fourth) layer of the backbone architecture (ResNet50-IBN) into a branch-based configuration.

\subsubsection{Bottleneck Transformer Branch (BoT)}
The attention branch module is designed to capture local dependencies through self-attention, recurring to the architectural modification from BoTnet~\cite{srinivas2021bottleneck}, also adopted and modified by \cite{article}. This modification replaces the last three bottleneck blocks of the fourth ResNet layer~\cite{he2016deep} with multi-head self-attention (MHSA) blocks (see Fig. \ref{fig:diag_b_branches}) to enable \textit{all2all} attention over 2D feature maps. The MHSA is applied with four heads following \cite{srinivas2021bottleneck}. 
The BoT block reduces parameters compared to the original stage while obtaining distinct information from the image through self-attention. 

\begin{figure*}[htb!]
\centerline{{\includegraphics[width=1.0\linewidth]{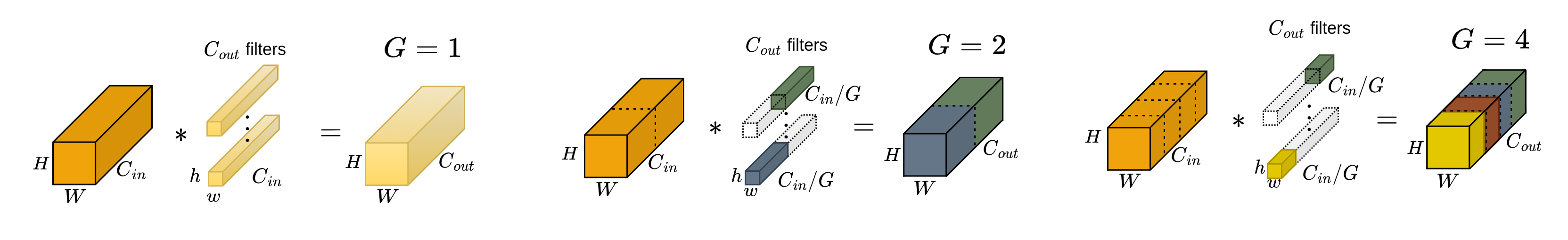}}}
\caption{Left: an example of the standard convolution. Middle: grouped convolution with $G=2$. Right: a grouped convolution with $G=4$.}
\label{fig:groupconv}
\end{figure*}

\subsection{Branch Grouped Convolution}
Focusing on the multi-branch architectural representation with loss splitting, we devise a lightweight alternative employing grouped convolutions \cite{krizhevsky2017imagenet}, also used in ResNeXt by \cite{xie2017aggregated}. 

Instead of performing convolutions over all input channels, the convolution occurs in $G$ groups. This method enables to generate an output embedding $f_N \in \re^\frac{D}{G}$, where $D$ is the output dimension of the concatenated output responses of $G$ groups with $C_{in}/G \times C_{out}/G$ filters, each corresponding to different branch.
Given that each branch computes $C_{out}$ by a fraction of $C_{in}$ it reduces the parameters count at that stage by a factor of $G$\footnote{By adopting branch grouped convolutions, \textit{e.g.} in the 2-branch, a reduction of 7M parameters is achieved when compared to the original baseline (1-Branch), as the last layer's parameter count is halved.}, as depicted in Fig. \ref{fig:groupconv}. 
Each branch $g$ operates at a different input feature group of channels $f_{L_3}(g) \in \mathbb{R}^{16 \times 16 \times \frac{1024}{G}}$, where $g=1,\cdots, G$ in contrast to the multi-branch where each branch acts on the same set of features $f_{L_3} \in \mathbb{R}^{16 \times 16 \times 1024}$.

This strategy allows to preserve the LBS architecture, where we generate $G$ separated features due to $G$ groups mimicking the $N$ branches, also preserving shared shallower layer parameters while deeper layers learning more complex features concerning each loss. Contrary to the inclusion of multiple $N$ branches that introduce excessive parameter overhead, the addition of extra branches in this scenario actually leads to a reduction of the model's complexity. We define these architectures as MBR-G and the training strategy remains the same as for the equivalent MBR-B.

\subsection{Leveraging Additional Information (LAI)}
We designed a CNN based strategy to take advantage of metadata inherent to the image capture. 
Unlike details such as vehicle type, model, or color, which must be inferred and may lead to erroneous results, specific camera and orientation metadata is easily accessible in real-world situations.
Following \cite{he2021transreid}, we zero initialize a side embedding matrix $A \in \re^{N \times D \times N_{cam} \times N_{view}}$, where $N_{cam} \times N_{view}$ represents the total number of available labels for cameras and view orientations.
However, opposed to \cite{he2021transreid} that computes positional embeddings at the patch-level, we add this information to each $f_N$ global embedding output. Depending on the input camera $c$ and view $v$ configuration, the corresponding side embeddings $A_{cv} \in \re^{N \times D}$ are selected and added to the global branches. Since the total number of patches of previous approach is significantly higher than the number of branches (256 vs 4), our approach requires significantly fewer parameters for the operation (\textit{e.g.}, $0.33M \times N$ vs $31.46M$ in Veri-776). 
On versions relying on grouped convolution we have 0.33M parameters since $f_g \in \mathbb{R}^{D}$ remains unchanged. 
These learnable parameters are added into the aggregated feature according to the image's camera and view configuration, as shown in Fig.\ref{fig:diagrama}, and are estimated during model training.

\subsection{Loss Branches and Training}
For training, a combination of classification loss and metric loss is used. Following the Loss-Branch-Split (LBS) strategy, each architecture's losses originates a dedicated branch per loss that are trained independently.

\begin{figure*}[hbt!]
\centerline{{\includegraphics[width=1.0\linewidth]{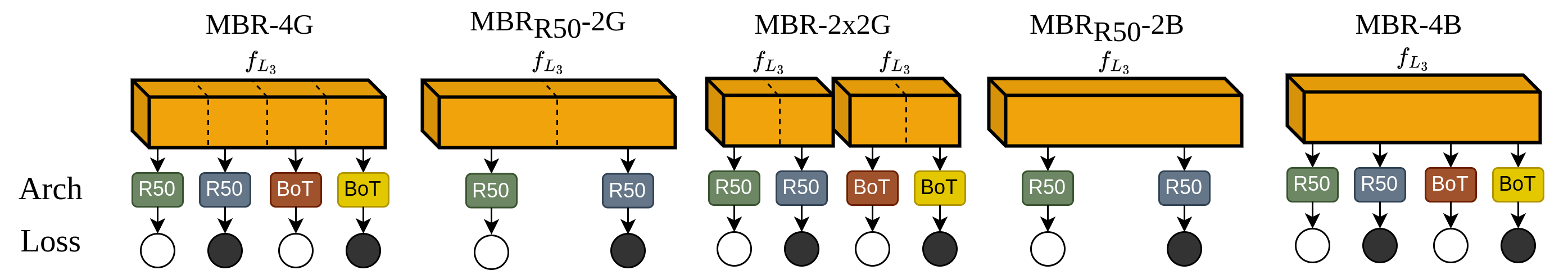}}}

\caption{MBR architectures variants. Feature maps are extracted from Layer 3 ($f_{L_3}$) and fed to different architectures and losses. (1) MBR-$\#$G refers to branching with $G$ groups, (2) MBR-$\#$B refers to branching by expansion, where entire $f_{L_3}$ is fed to $B$ branches and (3) MBR-2x2G refers to expanding features into two branches that operate two groups. Losses are represented by circles, where $\circ$ CE and $\bullet$ Triplet. 
} 

\label{fig:arches_ablate}
\end{figure*}

\subsubsection{Classification Branch}
The classification branch of the architecture is devoted to a classification task, extracting fine-grained information that can identify the specific ID. After the embedding extraction, a dense layer reduces the vector to the number of classes in training ($C$) from $\mathbb{R}^{D/G}$ to $\mathbb{R}^{C}$. Next, a softmax is applied and the cross entropy loss with label smoothing following \cite{szegedy2016rethinking} is computed, which is defined as:
\begin{equation}
L_{cls} = - \sum_{c=1}^{C} y_{gt}^i \log{y_p^i},      
\end{equation}
where $y_p^i = \frac{\exp{z^i}}{\sum_{c=1}^C \exp{z_j}}$ represents the softmax of the predicted class logits and $C$ the number of classes in training. $y_{gt}^i$ is the ground-truth label for the $i^{th}$ class, which is formulated with label smoothing as:
\begin{equation}
    y_{gt}^i =
    \begin{cases}
      1 - \frac{C-1}{C}\epsilon & \text{if $i = class$}\\
      \frac{\epsilon}{C} & \text{if $i \neq class$},\\
    \end{cases} 
\end{equation}
where $\epsilon$ is the smoothing parameter.

\subsubsection{Metric Branch}
The metric branch is responsible for capturing an embedding specialized on details representative from distinctive views. It uses the batch hard triplet loss \cite{hermans2017defense}, where batches are built using a random PK sampling, assembled with $P$ identities and $K$ samples belonging to an identity. Batch hard means that triplets are formed within each batch, by considering each image as anchor and obtaining the "hardest" samples, in a nutshell the hardest positive is the image of same identity with maximum distance from anchor, the hardest negative is an image from different identity closest to the anchor. This technique achieves better results than the conventional triplet approach because it mines hard triplets instead of choosing random positive and negative samples that can be uninformative by not producing any loss, as explained in \cite{hermans2017defense}. The loss is represented as:
\begin{equation}
    L_{tri} = \frac{1}{PK} \sum_{i=1}^{PK} \max[0, m + \max_{p \in P(a)} d(x_a,x_p) - \min_{n \in N(a)} d(x_a,x_n)],
\label{eqn:hardtripletloss}
\end{equation}
where $P$, $K$, $m$, $P(a)$, $N(a)$ are respectively the number of identities, samples belonging to that identity, distance margin threshold, positive and negative sample sets with respect to each given anchor in the batch. The $x_a$, $x_p$, $x_n$ represent the extracted embeddings of anchor, positive and negative samples. Moreover, $d()$ denotes the euclidean distance.

\subsubsection{Network Training}

Following the proposed LBS architecture, each branch loss is trained independently and the global objective loss $\mathcal{L}$ is obtained by means of weighted linear combination of each branch loss $L^i$, resulting

\begin{equation}
    \mathcal{L} = \sum_{i=1}^{N} \omega_{cls}^i L_{cls}^i +  \omega_{tri}^i L_{tri}^i,
    \begin{cases}
      \omega_{tri}^i=0  \text{ if } \text{Cls Branch} \\
      \omega_{cls}^i=0  \text{ if } \text{Metric Branch} \\
    \end{cases}
    \label{eq:loss_global}
\end{equation}

Each loss weight $\omega_i$ is tuned by cross-validation.

\section{Experiments, Results and Discussion}
In this section, we present our experimental results on two widely used datasets: Veri-776 \cite{liu2016deep} and Veri-Wild \cite{lou2019veri}. We begin with an overview of the datasets and their specifications, followed by a description of our proposed architecture implementation details. We report our findings in terms of percentage of mAP (Mean Average Precision) and CMC (Cumulative Match Curve) at rank 1 and 5, using the standard image-to-image retrieval process of ranking the entire gallery against each query image based on similarity scores. We also present ablation studies to consolidate our architectural proposals. Lastly, we compare our results with state-of-the-art methods. All reported values exclude post-processing techniques.

\subsection{Dataset}

The datasets we evaluate are Veri776~\cite{liu2016deep} and Veri-Wild~\cite{lou2019veri}. 
Veri-776 contains images of 776 vehicles captured by 20 different cameras with no view restrictions. The dataset is split into 37778 training images, 1678 query images, and 11579 gallery images.
Veri-Wild is composed of 40671 vehicles captured by 174 cameras with more restricted viewpoints (front, rear), but with challenging conditions of severe illumination and weather. The training set consists of 277597 images of 30671 identities, and the validation set contains three subsets (S, M, L) of 3000, 5000, and 10000 query images, and 38861, 64389, and 128517 gallery images, respectively.

\subsection{Implementation Details}

The baseline architecture and methods are implemented based on \cite{luo2019bag}. Instead of a ResNet50, we use a ResNet50-IBN \cite{pan2018two} given its effectiveness in ReID tasks and modified the last convolutional stage stride to 1.
Input images are resized to $256\times256$ pixels and augmented during training with standard augmentation techniques, such as $256\times256$ random crops from 10-pixel padded images, horizontal flips, and random erasing. 
To construct the training batches, we perform a PK sampling strategy similar to \cite{zhao2021heterogeneous,sun2020CFVMNet, he2020fastreid, article}, which is based on the number of images per identity for each dataset. On Veri-776 $P=6$ and $K=8$ and on Veri-Wild $P=32$ and $K=4$, resulting in batch sizes of $48$ and $128$, respectively.

All the model layers, apart from the BoT block, use pre-trained weights on ImageNet. The models are trained on each dataset during 120 epochs with an Adam optimizer using a learning rate of 1e-4, a linear warm-up of 10 epochs and decrement steps at epochs 40, 70, and 100 with a factor of $0.1$.
To align the randomly initialised BoT blocks with the pre-trained weights, we followed a two-step fine-tuning approach. Initially, we froze the first three backbone stages in architectures containing BoT, while training the remaining layers using the same optimizer with fixed learning rate of 1e-4 for 10 epochs. Subsequently, we conducted training for the complete model with the previous settings.
Classification loss has a label smoothing factor $\epsilon = 0.1$ and the triplet loss a margin $m=0.1$. The MBR hyper-parameters were empirically tuned through cross-validation, being the classification loss weights set to $\omega_{cls} =0.6$ and the metric loss weights set to $\omega_{tri} =1.0$.
On LAI models we used the metadata available in each dataset, \textit{i.e.}, the camera ID and vehicle view on Veri-776 and camera ID on Veri-Wild.

\subsection{Ablation Studies}

To demonstrate how the performance of V-ReID is influenced by increasing feature diversity through multiple branches, losses, and architectures, a series of ablation studies were conducted on the Veri-776 dataset. 
Fig.~\ref{fig:arches_ablate} illustrates some of the multiple branch architectures (MBR) that combines diverse losses and architectures, where $B$ denotes the number of features splited branches and $G$ denotes the number of branch grouped convolution splits. 

The architectures with convolutional blocks (R50) and Hybrid blocks (R50+BoT) are trained with combined losses (CE+Triplet), and their respective Loss-Branch-Split (LBS) variants, denoted as MBR\textsubscript{R50} and MBR, respectively.

\begin{table}[tb]
\centering
\caption{Ablation of the R50 variants in Veri-776.}
\label{tab:r50_ablation}
\resizebox{\columnwidth}{!}{
\begin{tabular}{c|cc|cccc|cc}
\hline
\textbf{Method}    & \textbf{LBS} & $\mathbf{G}\backslash\mathbf{B}$ & \textbf{Params} & \textbf{FLOPs} & \makecell{\textbf{Dim} \\ \boldmath{$f_{L3}(g)$}}  & \makecell{\textbf{Dim} \\ \boldmath{$f_{g}$}} &  \textbf{mAP}            & \textbf{CMC1}          \\ \hline 
R50      &  & 1$\backslash$ 1  & 23.5M  & 8.1G &1024        & 2048                      & 81.15          & 96.96          \\ \hline
R50-4G   & & \multirow{2}{*}{4$\backslash$1}    & \multirow{2}{*}{12M}  & \multirow{2}{*}{5.3G}  & \multirow{2}{*}{256}         & \multirow{2}{*}{2048}                & 82.81         & 97.38       \\
MBR\textsubscript{R50}-4G    &            \checkmark   &  &        &     &             &  & 82.47          & 96.84     \\ \hline
R50-2G   & & \multirow{2}{*}{2$\backslash$1}  & \multirow{2}{*}{16M} & \multirow{2}{*}{6.2G}   & \multirow{2}{*}{512}         & \multirow{2}{*}{2048}               & 83.04          & 97.14         \\
MBR\textsubscript{R50}-2G          &            \checkmark &   &    &         &             &  & 83.26          & 97.02     \\ \hline
R50-2x2G &  & \multirow{2}{*}{2$\backslash$ 2} &  \multirow{2}{*}{23.5M} & \multirow{2}{*}{8.1G} & \multirow{2}{*}{512}         & \multirow{2}{*}{4096}                   & 83.67            & 97.32         \\
MBR\textsubscript{R50}-2x2G         &             \checkmark  &   &       &      &             &  & \textbf{84.22} & 97.02     \\ \hline
R50-2B   & & \multirow{2}{*}{1$\backslash$ 2}  & \multirow{2}{*}{38.5M} & \multirow{2}{*}{11.9G} & \multirow{2}{*}{1024}         & \multirow{2}{*}{4096}                 & 81.82           & 96.96      \\
MBR\textsubscript{R50}-2B         &           \checkmark   &  & &             &             &  & 83.67          & \textbf{97.50}    \\ \hline
R50-4B   &  & \multirow{2}{*}{1$\backslash$4}  & \multirow{2}{*}{69.6M} & \multirow{2}{*}{19.6G}  & \multirow{2}{*}{1024}        & \multirow{2}{*}{8192}                  & 82.31          & 97.32     \\
MBR\textsubscript{R50}-4B     &         \checkmark  &       &        &     &             &  & 83.89          & \textbf{97.50}  \\ \hline
\end{tabular}
}
\end{table}

\subsubsection{Diversity with Multiple-branches}

Firstly, we evaluate how to generate feature diversity with R50 blocks. In Table~\ref{tab:r50_ablation}, we present the results for different architectures, which can be categorized into branching by expansion (R50-2B, R50-4B), branching by grouping (R50-4G, R50-2G), or a combination of both (R50-2x2G).

Results in Table~\ref{tab:r50_ablation} reveal the following remarks: 
(i) architecture branching always results in mAP gains; (ii) branching by grouping achieves improved performance compared with branching by expansion. Without LBS, branching by expansion leads to replicating identical branches, relying on the same features to learn identical tasks. In contrast, grouping provides each branch with a distinct set of channel-grouped features to perform the same task; (iii) combining the two sets of groups (R50-2x2G) achieves a boost of $2.52\%$ mAP and $0.36\%$ CMC1, with on-par parameters to baseline architecture, and a boost of $1.66\%$ mAP and $0.42\%$ CMC1 with half parameters (R50-4G).

\subsubsection{Diversity with Loss Branch Split}

The use of the LBS in all architectures reported in Table~\ref{tab:r50_ablation} globally improves both metrics. 
The MBR\textsubscript{R50}-2x2G architecture demonstrates a $0.55\%$ increase in mAP, while the MBR\textsubscript{R50}-2B and MBR\textsubscript{R50}-4B models exhibit significant mAP gains of $1.85\%$ and $1.58\%$, respectively. 
Except for MBR\textsubscript{R50}-4G, all architectures improved mAP scores with LBS compared to the metric and classification combined loss, typically with a trade-off between mAP and CMC1 scores.
These findings support the idea that multiple branches and LBS can introduce greater diversity to the feature representations for V-ReID.

\subsubsection{Diversity with Attention}

Table~\ref{tab:hybrid_ablate} presents the results of combining global and attention branches, including a baseline with BoT and three attention-based architectures: MBR-4G, MBR-2x2G, and MBR-4B. Overall we note a consistent trade-off behaviour between the metrics, as observed in previous evaluations. However, despite these overall lower scores, the combination of multiple branches with attention and LBS (MBR-4B) manages to surpass previous results across all models with $84.72\%$ mAP and $97.68\%$ CMC1. This indicates the potential of leveraging combined branches with BoT to achieve superior performance in V-ReID tasks.

\subsubsection{LAI}
An ablation analysis of the proposed LAI module is provided in Table~ \ref{tab:LAI}. Our experiments on all solutions reveal a consistent improvement around 1\% in mAP and a slight increases in CMC1.

\begin{table}[tb]
\centering
\caption{Ablation of Hybrid variants (R50+BoT) in Veri776.}
\label{tab:hybrid_ablate}
\resizebox{\columnwidth}{!}{%
\begin{tabular}{c|cc|cccc|cc}
\hline
\textbf{Method}  & \textbf{LBS} & $\mathbf{G}\backslash\mathbf{B}$ &\textbf{Params} & \textbf{FLOPs} & \makecell{\textbf{Dim} \\ \boldmath{$f_{L3}(g)$}}       & \makecell{\textbf{Dim} \\ \boldmath{$f_{g}$}}       & \textbf{mAP}            & \textbf{CMC1}          \\ \hline
BoT    &      &  1$\backslash$ 1   & 18.8M  & 7.2G & 1024                 & 2048          &               80.09          & 96.78       \\ \hline
Hybrid-4G   &   & \multirow{2}{*}{(2+2)$\backslash$ 1} & \multirow{2}{*}{11.7M}  & \multirow{2}{*}{5.2G} & \multirow{2}{*}{256}                  & \multirow{2}{*}{2048}                & 82.04          & 96.96        \\
MBR-4G          &     \checkmark     &        &            &          &               &  & 82.67          & 97.02         \\ \hline
Hybrid-2x2G &  & \multirow{2}{*}{      2$\backslash$ 2} & \multirow{2}{*}{18.8M} & \multirow{2}{*}{7.9G} & \multirow{2}{*}{512}                  & \multirow{2}{*}{4096}                        & 82.02          & 96.78        \\
MBR-2x2G        &      \checkmark    &        &        &              &               &  & 82.57          & 97.32          \\ \hline
Hybrid-4B   &   & \multirow{2}{*}{  1$\backslash$ \textnormal{(2+2)}} & \multirow{2}{*}{59.1M} & \multirow{2}{*}{17.8G} & \multirow{2}{*}{1024}                 & \multirow{2}{*}{8192}                       & 83.30          & 97.62          \\
MBR-4B       &  \checkmark     &         &  & \multicolumn{1}{l}{} &               &  & \textbf{84.72} & \textbf{97.68} \\ \hline
\end{tabular}%
}
\end{table}

\begin{table}[t]
    \caption{Ablation of using LAI on Veri-776 \cite{liu2016deep}. Values after $\uparrow$ or $\downarrow$ represent gains or losses over respective architectures.}
    \begin{center}
    \resizebox{\linewidth}{!}{
    \begin{tabular}{c|ccc}
    \hline
    \textbf{Method} & \textbf{mAP} & \textbf{CMC1} & \textbf{CMC5}  \\ \hline

    MBR-4G-LAI & 83.49 ($\uparrow$0.82)         & 97.02 ($\uparrow$0)         &  98.81 ($\uparrow$0.0)     \\ 
    MBR\textsubscript{R50}-2G-LAI & 84.00 ($\uparrow$0.75)         & 97.44 ($\uparrow$0.42)         & 99.01 ($\uparrow$0.5)      \\

    MBR\textsubscript{R50}-2B-LAI & 84.87 ($\uparrow$1.2)         & 97.56 ($\uparrow$0.06)          & 98.75 ($\downarrow$0.06)         \\ 

    MBR-4B-LAI & 85.63 ($\uparrow$0.91)         & 97.74 ($\uparrow$0.06)         & 99.05 ($\uparrow$0.24)  \\ \hline
    
    \end{tabular}}
    \label{tab:LAI}
    \end{center}
\end{table}

\begin{table*}[tb]
\begin{center}
\caption{Result comparison with SOTA on Veri-776 and Veri-Wild datasets. Note that * means the use of extra data.}
\resizebox{1.0\linewidth}{!}{
\begin{tabular}{c||c|c|c||c|c|c|c|c|c|c|c|c||c|c}
\hline
\multirow{3}{*}{\textbf{Method}} & \multicolumn{3}{c||}{\textbf{Veri-776} \cite{liu2016deep}} & \multicolumn{9}{c||}{\textbf{Veri-Wild} \cite{lou2019veri}} & \multirow{3}{*}{\textbf{Params [M]}} & \multirow{3}{*}{\textbf{Dim $f_g$}}  \\ \cline{2-13}
 & \multirow{2}{*}{mAP} & \multirow{2}{*}{CMC1} & \multirow{2}{*}{CMC5} & \multicolumn{3}{c|}{Small} & \multicolumn{3}{c|}{Medium} & \multicolumn{3}{c||}{Large} &     &  \\ \cline{5-13}
 & & & & mAP & CMC1 & CMC5 & mAP & CMC1 & CMC5 & mAP & CMC1 & CMC5 &   &  \\ \hline 
 
 \hline
PEVEN \cite{meng2020parsing}& 79.5 & 95.6 & 98.4 & 79.8 & 94.01 & 98.06 & 73.91 & 92.03 & 97.15 & 66.2 & 88.62 & 95.31 &  59.2 & 10240 \\ \hline 
SAVER \cite{khorramshahi2020devil}& 79.6 & 96.4 & 98.6 & 80.9 & 94.5 & 98.1 & 75.3 & 92.7 & 97.4 & 67.7 & 89.5 & 95.8 &  31 & \textbf{2048} \\ \hline 
GLAMOR \cite{suprem2020looking}& 80.34 & 96.53 & 98.62 & 77.15 & 92.13 & 97.43 & & & & & & &  38.5 & \textbf{2048} \\ \hline
CFVMNet \cite{sun2020CFVMNet}& 77.06 & 95.3 & 98.4 & & & & & & & & & & 38.5 & 6144 \\ \hline
FastREID \cite{he2020fastreid}& 81.9 & 97.0 & 98.9 & 85.37 & 95.68 & 98.96 & 80.48 & 94.52 & 98.23 & 73.65 & 91.91 & 96.84 &  23.5 & 2048 \\ \hline 
TransREID \cite{he2021transreid} *& 82.3 & 97.1 &  & 81.2 & 92.3 & 98.0 & & & & & & &  101 & 3840 \\ \hline 
HRCN \cite{zhao2021heterogeneous}& 83.1 & 97.3 & 98.9 & 85.2 & 94.0 &  & 80.0 & 91.6 &  & 72.2 & 88.0 &  &  55.4 & 3584 \\ \hline 
ANet \cite{quispe2021attributenet} & 81.2 & 96.8 & 98.4 & 85.8 & 95.9 & 99.0 & 81.0 & 94.5 & 98.1 & 73.9 & 91.6 & 96.7 & 67 & \textbf{2048} \\ \hline 
MUSP \cite{lee2022multiple} & 78.0 & 95.6 & 97.9 & & & & & & & & & & & \\ \hline
TANet \cite{article} & 80.5 & 95.4 & 98.4 & & & & & & & & & & 61.1 & 4096 \\ \hline
SSBVER \cite{khorramshahi2022scalable}& 82.1 & 97.1 & 98.4 & 82.64 & 95.11 & 98.53 & 77.49 & 93.37 & 97.45 & 70.09 & 90.14 & 95.67 &  23.5 & \textbf{2048} \\ \hline \hline

MBR-4G & 82.67 & 97.02 & 98.81 & 86.05 & 94.91 & 98.69 &  80.95 & 92.61 & 97.47 & 73.62 & 89.26 & 95.84 & \textbf{11.7} & \textbf{2048} \\ \hline 
MBR\textsubscript{R50}-2G & 83.25 & 97.02 & 98.51 & 86.04 & 95.31 & 98.73 & 81.15 & 93.63 & 97.79 & 73.94 & 90.18 & 96.1 & 16 & \textbf{2048} \\ \hline

 MBR\textsubscript{R50}-2x2G & 84.22 & 97.08 & \textbf{98.99} & 86.92 & 95.78 & 98.8 & 82.19 & 93.79 & 98.11 & 75.26 & 90.65 & 96.48 & 23.5 & 4096 \\ \hline
 MBR\textsubscript{R50}-2B  & 83.67 & 97.5 & 98.81 & 87.47 & 96.22 & 98.73 & 82.84 & 94.62 & 98.21 & 75.98 & 91.41 & 96.86 & 38.5 & 4096 \\ \hline 
MBR-4B & \textbf{84.72} & \textbf{97.68} & 98.81 & \textbf{87.97} & \textbf{96.29} & \textbf{99.06} & \textbf{83.46} & \textbf{95.16} & \textbf{98.27} & \textbf{77.15} & \textbf{92.28} & \textbf{97.14} & 59.1 & 8192 \\ \hline \hline

MBR-4B-LAI * & 85.63 & 97.74 & 99.05 & 88.12 & 96.29 & 98.93 & 83.81 & 95.18 & 98.28 & 77.41 & 92.48 & 97.17 & 60 & 8192 \\ \hline

\end{tabular}}
\label{tab:vreid_sota}
\end{center}
\end{table*}

\subsection{Results}
A comparison with state-of-the-art (SOTA) vehicle re-id is shown in Table \ref{tab:vreid_sota}. 

\subsubsection{Performance on Veri-776}

In this dataset our models outperform all other even without additional metadata. The MBR-4B model surpasses the best score on Veri-776 by 1.62\% mAP and 0.38\% CMC1 against HRCN \cite{zhao2021heterogeneous}. Also, our MBR\textsubscript{R50}-2G matches HRCN with $3.46\times$ less model parameters and a reduced embedding size. Adding the LAI module to our MBR-4B brings an additional gain of 0.91\% mAP and 0.06\% CMC1 surpassing TransREID \cite{he2021transreid} by 3.3\% mAP and 0.64\% CMC1.
Also given recent interest on model sizes and their throughput by \cite{khorramshahi2022scalable}, we point that our lightweight MBR-4G can surpass most older works and be very close to recent works with a fraction of the parameters.
\subsubsection{Performance on Veri-Wild} 
Our largest variant MBR-4B surpasses ANet \cite{quispe2021attributenet} by 2.17\% mAP and 0.39\% CMC1 on small set. Compared to FastREID \cite{he2020fastreid} trained with batch size 128 we obtain 2.6\% additional mAP and 0.61\% CMC1. 
The MBR\textsubscript{R50}-2G and MBR-4G architectures demonstrate competitive lightweight solutions, outperforming SSBVER~\cite{khorramshahi2022scalable} by at least 3.4\% mAP with less parameters. Additionally, they show superior performance compared to larger solutions such as HRCN~\cite{zhao2021heterogeneous} by 0.84\% mAP and 1.31\% CMC1.

\section{Conclusions}
The work developed exhibits competitive results in vehicle re-identification. Strength in diversity is critical for retrieval tasks such as V-ReID as comproved by the results. Performing the split with groups not only improves diversity with different inputs but also may be a major improve for future real-time applications given their reduction in model's size while maintaining performance. We hope future works may experiment diversified representation learning beneficial in retrieval tasks.

\bibliographystyle{./IEEEtran}
\bibliography{main}

\end{document}